# R-Theta Local Neighborhood Pattern for Unconstrained Facial Image Recognition and Retrieval

Soumendu Chakraborty*, Satish Kumar Singh, and Pavan Chakraborty

*Corresponding Author Email: soum.uit@gmail.com, Mobile: +91-9897034787

**Abstract—In this paper R-Theta Local Neighborhood Pattern (RTLNP) is proposed for facial image retrieval. RTLNP exploits relationships amongst the pixels in local neighborhood of the reference pixel at different angular and radial widths. The proposed encoding scheme divides the local neighborhood into sectors of equal angular width. These sectors are again divided into subsectors of two radial widths. Average grayscales values of these two subsectors are encoded to generate the micropatterns. Performance of the proposed descriptor has been evaluated and results are compared with the state of the art descriptors e.g. LBP, LTP, CSLBP, CSLTP, Sobel-LBP, LTCoP, LMeP, LDP, LTrP, MBLBP, BRINT and SLBP. The most challenging facial constrained and unconstrained databases, namely; AT&T, CARIA-Face-V5-Cropped, LFW, and Color FERET have been used for showing the efficiency of the proposed descriptor. Proposed descriptor is also tested on near infrared (NIR) face databases; CASIA NIR-VIS 2.0 and PolyU-NIRFD to explore its potential with respect to NIR facial images. Better retrieval rates of RTLNP as compared to the existing state of the art descriptors show the effectiveness of the descriptor.**

1. Introduction

Facial image analysis is an intrinsic problem of face recognition and retrieval. Several descriptors have been proposed to recognize facial images under varying illumination, expression, light, scale and pose. Initially some of the descriptors namely eigenface [1], fisherface [1], variations of PCA [2][3][4][5][6], and linear discriminant analysis (LDA) [7][8][9]were proposed for facial images taken in controlled environment. These descriptors are unable to handle the variations due to illumination, expression, pose, light, scale, and partial occlusion in unconstrained environment. In recent literatures, deep learning based image descriptors have been proposed where a Convolutional Neural Network (CNN) is trained to classify the facial images [32-33]. The proposed descriptor is a hand-crafted descriptor and does not require any learning (training) mechanism to represent images (single or multiple) of a given class. Apart from this, the characteristic of the proposed descriptor is very much different from VGG face recognition model that make use of Convolutional Neural Networks (CNNs) to extract the features from an image. One of the most commonly used descriptor; local binary pattern (LBP) [10] captures the relationship amongst the pixels in the local neighborhood of the reference pixel. Although the LBP is a rotation invariant descriptor, but fails while tested on complex facial datasets (i.e. sever changes in illumination, pose, light, scale, and expression etc.). Another local descriptor namely, Center Symmetric Local Binary Pattern (CSLBP) [11][12] modifies LBP to reduce the length of the descriptor so as to improve the performance in terms of the time and accuracy of LBP in region based image matching. Sobel Local Binary Pattern (Sobel-LBP) [27] an advancement over LBP, enhances the edges in an image with the sobel operator and LBP encodes the circular gradients of an image. Performance of LBP and CSLBP significantly degrades under lighting variations [13]. A number of preprocessing techniques have been employed by Tan et al. in [13] to enhance the quality of the images taken under sever lighting variations. Local Ternary Pattern (LTP) proposed by Tan et al. defines a zone with a threshold. Grayscales falling within this zone are encoded as zeros, whereas above and below the zone as +1 and -1 respectively. Performance of LTP is better than LBP and CSLBP under uniform lighting variations. However, under unconstrained illumination, light, scale, and pose variations the LTP fails to achieve the comparable recognition accuracy. Center Symmetric Local Ternary Pattern (CSLTP) [14] captures local gradient information and performs well under illumination variations. Binary Rotation Invariant and Noise Tolerant (BRINT) [30] descriptor computes averages of pixels on different radii and LBP is computed on these average intensities. It has been illustrated that, BRINT is robust to illumination variations, rotation changes, and noise [30]. Multi-Block LBP (MBLBP) computes average values of intensities using $2 \times 3$ mask, over which LBP is computed [31]. Multi-Block LBP achieves 8% improvement over original LBP feature [31]. Local Mesh Pattern (LMeP) [29] similar to LBP, encodes the relationships amongst the pixels in the local neighborhood of the center pixel. Semi Local Binary Pattern (SLBP) proposed in [15] computes region wise average values of grayscales and encodes the relationship of these average grayscales to generate the micropatterns. SLBP achieves better accuracy than LBP under noise and global illumination variations. There exist some other descriptors such as Local Derivative Pattern (LDP) [16], Local Ternary Co-occurrence Pattern (LTCoP) [28], and Local Tetra Pattern (LTrP) [17] which computes the features in higher order derivative planes in different directions. These descriptors perform well under controlled illumination, pose, and expression variations. However, lengths of these descriptors are large causing non suitability for real time applications for big data sets. The recognition accuracy of derivative based local descriptors under unconstrained environment is not comparable while applying on the data set under constrained environment.

The organization of the rest of the paper is as follows. Section 2 elaborates the motivation and proposition of the descriptor. Various performance measures to evaluate the effectiveness of the proposed descriptor are given in section 3. Various experiments have been performed and the obtained results are compared with the state of the art descriptors in section 4. A brief discussion on the performance of the proposed descriptor on NIR facial images is given in section 5. The work reported through this paper is concluded in section 6.

## 2. Proposed r-theta local neighborhood pattern

### 2.1. Motivation

Distinctive information that exists in a facial image tends to spread around the local region of the reference pixel. Most of the descriptors confine the local region to the pixels on the circumference of a circle ignoring the pixels within the circular region. If we try to capture the relationship of the reference pixel with the pixels within the circular neighborhood then the length of the micropattern increases beyond the acceptable limits. Motivated from the existing local descriptors like LBP, LTP, LDP and so forth, we propose a descriptor which not only captures additional distinctive information that exists in the local region but also maintains the length within the acceptable limits. Result analysis confer that these additional distinctive relationships are useful enough to significantly increase the accuracy of the proposed descriptor under constrained and unconstrained environment.

### 2.2. Proposed descriptor

Local pattern descriptors try to capture local relationships amongst the pixels and encode it into binary codes which effectively recognize the facial image under illumination, pose and expression variations. Proposed R-theta local neighborhood pattern (RTLNP) captures relationship amongst the pixels across angular as well as radial widths.

Fig. 1. shows the template of the local neighborhood of the reference pixel $G_0$. The reference pixel is the

Published in "Multimedia Tools and Applications, vol-78, no-11, pp. 14799-14822, DOI: 10.1007/s11042-018-6846-z, (2019). (Springer) ISSN/ISBN: 1573-772"

current pixel being encoded shown in yellow in Fig. 1. Radius $R_n = n | n = 1, 2 \ldots$ denotes the distance of the neighboring pixel from $G_0$ ($R_0 = 0$, denotes the distance of $G_0$ from $G_0$). $G_{1,R_1}$ denotes the first neighboring pixel point of $G_0$ at $R_1$, $G_{2,R_1}$ denotes the second neighboring pixel point of $G_0$ at $R_1$. Similarly $G_{l,R_n}$ denotes the $l^{th}$ ($l$ is defined in (3)) neighboring pixel point of $G_0$ at $R_n$. Intensity of a pixel point $G_{l,R_n}$ is denoted as $I(G_{l,R_n})$. Proposed descriptor divides the local neighborhood of the reference pixel $G_0$ into sectors of equal angular width $\Delta\theta$ as shown in Fig. 2. Each sector is then divided into two subsectors of radial widths $R_{in}$ and $R_{out} - R_{in}$.

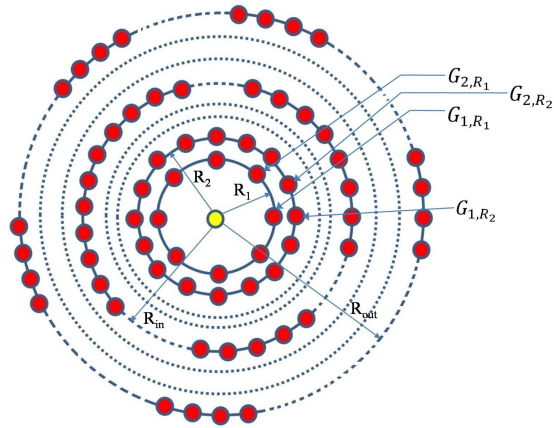

Fig. 1. Template showing the local neighborhood of the reference pixel in yellow. Neighboring pixels are shown in red.

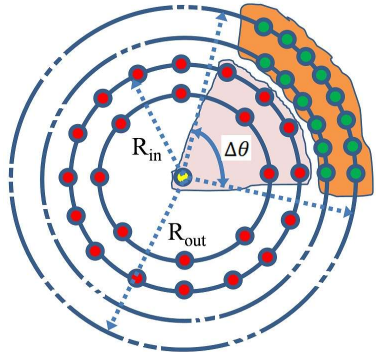

Fig. 2. Template showing angular and radial width of the sectors and subsectors in the local neighborhood of the reference pixel.

Inner and outer subsectors are shown in light pink and orange regions respectively in Fig. 2.

The number of sectors in the local neighborhood of the reference pixel is computed as

$$S = \lfloor 360/\Delta\theta \rfloor \qquad (1)$$

Where $\theta = k\Delta\theta + \theta_0 | k = 0,1,2..S$ and $\theta_0 = 0°$

The number of neighbors at given radius $R_n$ of a sector can be defined as

$$N_n^s = \left\lceil \frac{N_n}{S} \right\rceil = \left\lceil \frac{R_n \Delta\theta}{45} \right\rceil \qquad (2)$$

Where $N_n = 8R_n$ is the total number of pixels in the local neighborhood of the reference pixel at $R_n$. The radius $R_n$ is defined as $R_n = n$.

The index of the $k^{th}$ neighbor in the $j^{th}$ sector at radius $R_n$ is denoted as $l_{k,j}^{R_n}$ and defined as

$$l_{k,j}^{R_n} = \begin{cases} 8R_n, & if \left(\left(\left\lceil \frac{R_n \Delta\theta (j-1)}{45} \right\rceil + k\right) > 8R_n\right) \\ \left\lceil \frac{R_n \Delta\theta (j-1)}{45} \right\rceil + k, & otherwise \end{cases} \quad (3)$$

The proposed method precisely index each pixel in the local neighborhood. Each pixel in the inner and outer radius are indexed using (3). Pixels of the subsectors for $R_{in}$ and $\Delta\theta$ are indexed as $I(G_{l_{k,j}^{R_n}, R_n})$ where $l_{k,j}^{R_n} = \left\lceil \frac{R_n \Delta\theta (j-1)}{45} \right\rceil + k$, and $R_n = n | n = 1, .. in$. The subsector is denoted by $j$ and $k$ represents the $k^{th}$ neighbor on $R_n$ of the $j^{th}$ sector ($j = 1,2,..S$, where $S$ is the total number of sectors). Hence the $k^{th}$ neighbor on $R_1 = 1$ of $j = 1$ for $\Delta\theta = 36°$ is $I\left(G_{\left\lceil \frac{1 \times 36 \times 0}{45} \right\rceil + k, 1}\right) = I(G_{1,1})$ where $k = 1,2..\left\lceil \frac{R_n \Delta\theta}{45} \right\rceil$ (please note that $\left\lceil \frac{R_n \Delta\theta}{45} \right\rceil = 1$ for $R_1 = 1$ and $\Delta\theta = 36°$). Similarly $k^{th}$ neighbors on $R_1 = 1$ of $j = 2$ is $I\left(G_{\left\lceil \frac{1 \times 36 \times 1}{45} \right\rceil + k, 1}\right) = I(G_{2,1})$ as $k = 1$. Pixels at $R_2 = 2$, $j = 1$ and $k = 1,2..\left\lceil \frac{R_n \Delta\theta}{45} \right\rceil$ are indexed as $I(G_{1,2})$ and $I(G_{2,2})$ (please note that $\left\lceil \frac{R_n \Delta\theta}{45} \right\rceil = 2$). Similarly pixels at $R_2 = 2$, $j = 2$ and $k = 1,2..\left\lceil \frac{R_n \Delta\theta}{45} \right\rceil$ are indexed as $I(G_{3,2})$ and $I(G_{4,2})$ and so on. Pixels for the outer subsectors with $R_{in} = 3$ and $R_{out} = 6$ are indexed using $R_n = n | n = (in + 1), (in + 2).. out$. The $k^{th}$ neighbor on $R_4 = 4$ of $j = 1$ is indexed as $I(G_{k,4})$. Hence all the neighbors on $R_4 = 4$ of $j = 1$ are indexed as $I(G_{1,4})$, $I(G_{2,4})$, $I(G_{3,4})$ and $I(G_{4,4})$. Similarly $k^{th}$ neighbor on $R_4 = 4$ of $j = 2$ is indexed as $I(G_{4+k,4})$. Hence all the neighbors on $R_4 = 4$ of $j = 2$ are indexed as $I(G_{5,4})$, $I(G_{6,4})$, $I(G_{7,4})$ and $I(G_{8,4})$. If the number of sectors is more than the number of pixels at a particular radius then $l_{k,j}^{R_n}$ is bounded by $8R_n$. So, the pixel on $R_1 = 1$ of $j = 10$, for $\Delta\theta = 36°$ and $k = 1$ is indexed as $I(G_{8,1})$. Hence it is evident that, all the pixels of inner subsectors (defined from $R_1 = 1$ to $R_{in} = in$) and outer subsectors (defined from $R_{in+1} = in + 1$ to $R_{out} = out$) of sectors ($j = 1,2,..S$, where $S$ is the total number of sectors) are indexed correctly.

The average intensity of the pixels within the $j^{th}$ inner subsector is computed as

$$A_j^{in} = \left\lceil \frac{\left(\sum_{n=0}^{in} \sum_{k=1}^{N_n^S} I(G_{l_{k,j}^{R_n}, R_n})\right)}{\sum_{n=0}^{in} N_n^S} \right\rceil \quad (4)$$

Similarly average intensity of the pixels within the $j^{th}$ outer subsector is computed as

$$A_j^{out} = \left[ \frac{\sum_{n=in+1}^{out} \sum_{k=1}^{N_n^S} I(G_{l_{k,j}^{R_n},R_n})}{\sum_{n=in+}^{out} N_n^S} \right] \quad (5)$$

The RTLNP is defined using these average values computed from the inner and outer subsectors as

$$RTLNP_{R_{in},R_{out},\Delta\theta}(S) = \sum_{j=1}^{S} 2^{j-1} \, C(A_j^{in}, A_j^{out}) \quad (6)$$

The encoding function $C(.)$ is defined as

$$C(A_s^{in}, A_s^{out}) = \begin{cases} 0, & if \, A_s^{in} \leq A_s^{out} \\ 1, & else \end{cases} \quad (7)$$

Finally histogram of $RTLNP$ is computed, which represents the facial image as a feature vector. $\chi^2$ distance [13] is used to measure the similarity between two histograms, as it performs better than other measures (Euclidian, cosine, emd, Canberra, L1, and D1 [17]) in the experiment (Please refer Fig. 5.). Similarity measure $S_{\chi^2}(.,.)$ is defined as

$$S_{\chi^2}(X,Y) = \frac{1}{2} \sum_{i=0}^{q} \frac{(x_i - y_i)^2}{(x_i + y_i)} \quad (8)$$

where $S_{\chi^2}(X,Y)$ is the $\chi^2$ distance computed on two vectors $X = (x_1, \ldots, x_q)$ and $Y = (y_1, \ldots, y_q)$. Nearest one neighbor (1 NN) classifier is used as reported in [16] to compute the minimum $\chi^2$ distance between the probe and the gallery images. As similar regions of the probe and gallery images are effectively identified by 1NN classifier with optimal computational cost [16].

A sample image and the circular neighborhood have been shown in Fig. 3. Yellow and red dots in Fig. 3 show the reference pixel and the neighboring pixels respectively. The sample image has been encoded to generate $RTLNP$ for $R_{in} = 2$, $R_{out} = 3$ and $\Delta\theta = 45°$. Light pink and orange sub regions in Fig. 3 show the inner and outer subsectors of the first sector respectively. Average values of the respective inner and outer subsectors are shown by light pink and orange dots respectively. In Fig.3. the region of the reference pixel is defined for a sector j = 1,2..S, $R_{in} = 2$, $R_{out} = 3$ and $\Delta\theta = 45°$, $S = \lfloor 360/\Delta\theta \rfloor = 8$ and the inner subsector varies from $R_1$ to $R_{in}$ and the outer subsector varies from $R_{in+1}$ to $R_{out}$. Please note that the reference pixel (i.e. pixel at $R_0$) is explicitly added to the inner sector sum.

Encoding function as shown in (7) compares the average values of the inner and outer subsectors to generate the binary codes. For example $C(166,164)$ generates binary 1 and $C(154,161)$ generates binary 0. The size of the micropattern (number of bits) generated by $RTLNP$ depends upon the angular width $\Delta\theta$. In this example for $\Delta\theta = 45°$ the size of the micropattern is 8 bit long.

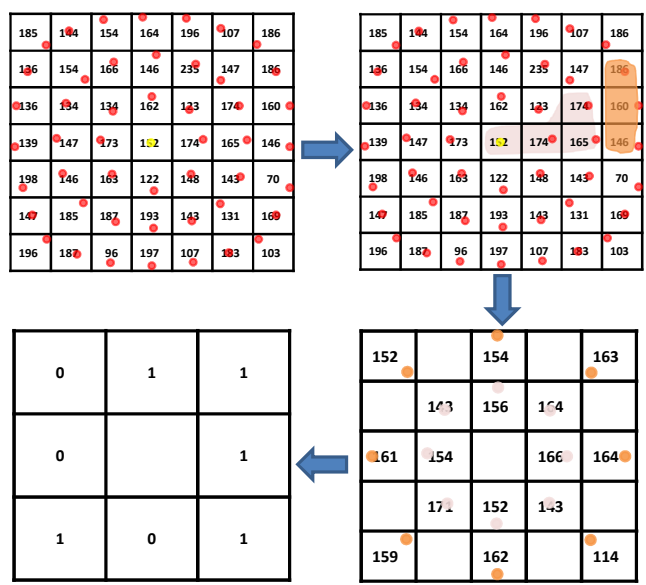

Fig. 3. Illustration of RTLNP with $R_{in} = 2$, $R_{out} = 3$ and $\Delta\theta = 45°$.

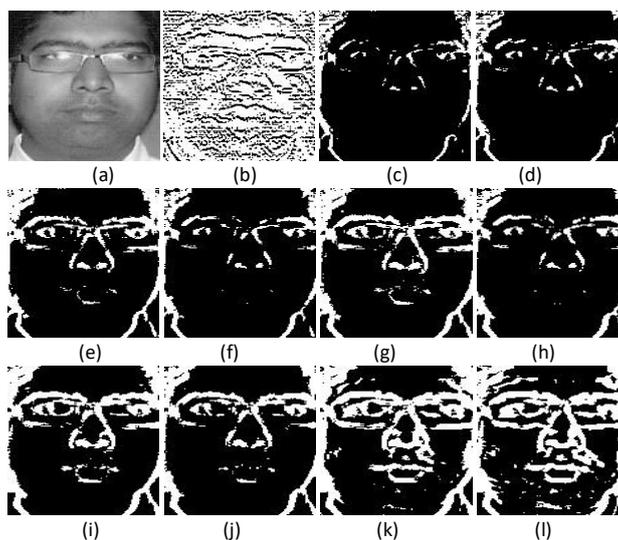

Fig. 4. Visual representation of RTLNP feature images: (a) original image, (b) image with $R_{in} = 1$, $R_{out} = 3$ and $\Delta\theta = 36°$, (c) image with $R_{in} = 2$, $R_{out} = 3$ and $\Delta\theta = 36°$, (d) image with $R_{in} = 2$, $R_{out} = 4$ and $\Delta\theta = 36°$, (e) image with $R_{in} = 3$, $R_{out} = 4$ and $\Delta\theta = 36°$, (f) image with $R_{in} = 2$, $R_{out} = 5$ and $\Delta\theta = 36°$, (g) image with $R_{in} = 3$, $R_{out} = 5$ and $\Delta\theta = 36°$, (h) image with $R_{in} = 2$, $R_{out} = 6$ and $\Delta\theta = 36°$, (i) image with $R_{in} = 3$, $R_{out} = 6$ and $\Delta\theta = 36°$, (j) image with $R_{in} = 3$, $R_{out} = 6$ and $\Delta\theta = 24°$, (k) image with $R_{in} = 3$, $R_{out} = 6$ and $\Delta\theta = 72°$, (l) image with $R_{in} = 3$, $R_{out} = 6$ and $\Delta\theta = 120°$.

Visual changes with changing $R_{in}$, $R_{out}$, and $\Delta\theta$ are shown in Fig. 4. If we carefully observe Fig. 4(d) and (e), it can be seen as the inner radius is increased, fine details around the lip region of the facial image are captured. Fine details become more prominent as the outer radius is increased in Fig. 4(e), (g), and (i). The effect of increasing $\Delta\theta$ has been shown Fig. (j), (k), and (l). As we increase $\Delta\theta$ more and more distinctive relationships are captured, which increases intra-class dissimilarity. Hence the retrieval accuracy tends to reduce with increasing value of $\Delta\theta$. Suitable values of $R_{in}$, $R_{out}$, and $\Delta\theta$ increases the retrieval accuracy of the proposed descriptor RTLNP. The unique features of RTLNP can be summarized as follows

1) Proposed descriptor captures relationship amongst the pixels using inner and outer regions. Instead of taking the central pixel as the reference pixels, the proposed descriptor takes the inner region around the reference pixel within a particular sector as the reference.

2) The entire region is divided into sectors of equal angular width $\Delta\theta$, which helps in capturing the distinctive relationships that exists amongst the pixels of a part of the local region.

3) Average values within the inner and outer subsectors are the appropriate representatives of the local region which are used to encode the micropatterns.

4) Micropatterns encoded using RTLNP more accurately discriminates inter-class facial images taken in constrained and unconstrained environment.

5) Even though most of the pixels within a local region contribute in encoding the RTLNP the length of the descriptor is comparable to the length of the state of the art descriptors.

6) There is a major difference in the structure of the RTLNP as compared to BRINT [30]. The proposed descriptor divides the operator into variable sized regions to compute the averages. The binary codes are computed by comparing the averages of inner and outer regions, whereas BRINT divides the operator across various radii only and computes the average per radius. These averages are compared with the center pixel to generate the binary pattern. It must be noted that in RTLNP inner averages are also varying with the varying sectors, whereas the value of the center pixel is fixed for the BRINT operator. RTLNP produces only one pattern for a particular $R_{in}$, $R_{out}$, and $\Delta\theta$, whereas BRINT produces two micropatterns per radius, which are eventually concatenated while computing the feature vector.

7) Computational complexity of the proposed descriptor is comparable with the state of the are descriptors like MB-LBP [31]. The proposed descriptor computes indices of, the inner subsectors with $((2 + R_{in}^2) \times R_{in})$ operations and the outer subsectors with $\left((2 + (R_{out} - R_{in})^2) \times (R_{out} - R_{in})^2\right)$ operations as analyzed in II.D. It is apparent from the analysis that computationally the region wise (sector wise) indexing for a specific $R_{in}$, and $R_{out}$ is constant for the proposed descriptor and computationally comparable to the state of the art descriptors like MB-LBP.

Features at various scales are extracted by the proposed descriptor by varying the size of the inner and outer subsectors, which are similar to the multi-scale features extracted by the state of the art MB-LBP. The scale of the inner and outer subsectors is changed by varying $R_{in}$ and $R_{out}$ respectively. As shown in Fig.4(b) the proposed descriptor extracts dominant features along with some irrelevant features at $R_{in} = 1$. As shown in Fig.4(c) with increasing $R_{in}$ relevant features become more prominent and irrelevant features get eliminated. It is evident in Fig.4(e), Fig.4(g), and Fig.4(i) that relevant features become more prominent with increasing $R_{out}$. The scale of the RTLNP operator is also changed by varying angular width $\Delta\theta$. It have been demonstrated in Fig.4(i), Fig.4(j), Fig.4(k), and Fig.4(l) that the proposed descriptor extracts dominant features at $\Delta\theta = 36°$.

### 2.3. Feature image analysis under varying conditions

The robustness of the proposed descriptor under variations in illumination, pose and with self occlusion is due to the fact that it precisely identifies significant features of a face (eyes, nose, and lips). To illustrate how RTLNP extracts prominent features of a facial image, the feature images under illumination, pose and expression variations are computed and shown in Fig. 5. and Fig. 6. respectively. The feature images are computed with $R_{in} = 3$, $R_{out} = 6$ and $\Delta\theta = 36°$. Original images with varying illumination are shown in Fig. 5(a-c) and the corresponding feature images are shown in Fig. 5(d-f). It is evident from the feature images that even with 80% reduction in illumination the proposed descriptor is able extract the significant features. The original images shown in Fig. 6(a-d) are captured with variations in pose as well as illumination. The proposed descriptor extracts visible features such as left-eye, nose, and some portion of lips from the left profile as shown in Fig. 6(e). Similarly, relevant features are extracted by RTLNP from right, up and down profile images as shown in Fig. 6(f-h). As relevant significant features are effectively computed by RTLNP from the visible portion of a facial image, it achieves better results under pose and illumination variations.

To show that RTLNP computes features from the visible regions of a self occluded facial image with expression and illumination variations. The features are computed with occluded eye, lips and nose with expression and illumination variations as shown in Fig.7. Fig.7(a) shows the image with occluded left-eye, major portion of the nose and some portion of lips. The corresponding feature image shown in Fig. 7(d) illustrates that the descriptor computes visible features correctly from image which effectively reduces the intra-class dissimilarity. As we reduce the occluded region (as shown in Fig. 7(b-c)) the descriptor extracts more and more relevant features (as shown in Fig.7(e-f)).

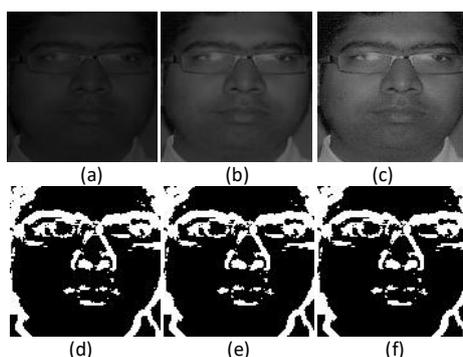

Fig. 5. Visual representation of RTLNP feature images for varying illumination: (a) original image with 20% illumination, (b) original image with 40% illumination, (c) original image with 60% illumination, (d) Feature image with 20% illumination, (e) Feature image with 40% illumination, (f) Feature image with 60% illumination.

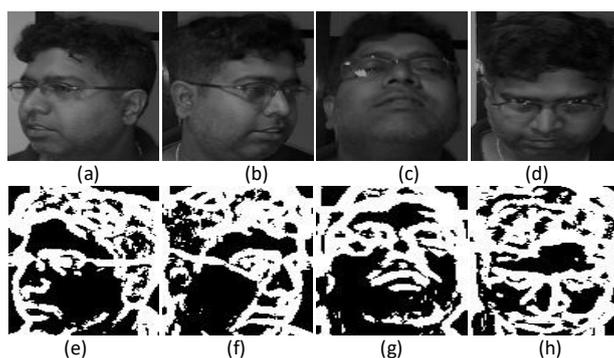

Fig. 6. Visual representation of RTLNP feature images for varying pose: (a) original image left profile, (b) original image right profile, (c) original image up profile, (d) original image down profile (e) Feature image left profile, (f) Feature image right profile, (g) Feature image up profile, (h) Feature image down profile.

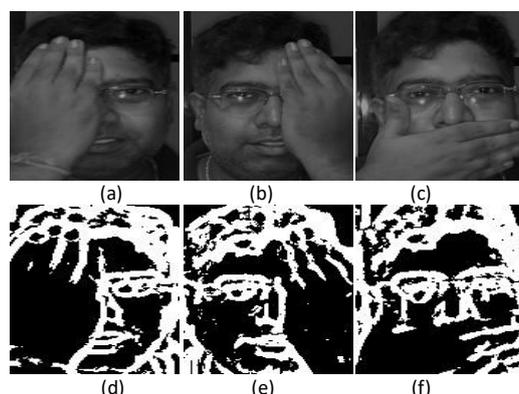

Fig. 7. Visual representation of RTLNP feature images with self occlusion: (a) original image with occluded right eye and some portion of the lips, (b) original image with occluded left eye and some portion of the lips, (c) original image with occluded lips, (d) Feature image with occluded right eye and some portion of the lips, (e) Feature image with occluded left eye and some portion of the lips, (f) Feature image with occluded lips.

The complementary information extracted by RTLNP from visible regions of a facial image are relevant features of a face. Hence the performance of the proposed RTLNP is better under illumination, pose, and expression variations with occlusion.

*2.4. Complexity analysis of the proposed descriptor*

The proposed method computes number of sectors by dividing the angular width of the local neighborhood in equal parts, this division is independent of the image size. The number of neighbors at a particular radius is computed by one division and one multiplication. Total number of operations required to compute the number of neighbors in the inner sector is $2 \times R_{in}$, where "in" is the index for the inner subsector. Similarly, number of operations required to compute index of the $k^{th}$ neighbor per sector at radius $R_n$ is $((2 + k) \times R_{in})$ (i.e. one division, one multiplication and k summations). Now k varies from 1 to the number of neighbors per sector $N_s^n$, which is dependent on $R_n = n$, where n varies from 1 to $in$. Hence total number of operations required to compute index for the inner subsectors is $((2 + R_{in}^2) \times R_{in})$. Similarly, number of operations required to compute the inner subsector average per sector is $(R_{in}^3 + R_{in}^2 + 1)$ which includes one division. Hence total number of operations for inner subsector are $\left((2 \times R_{in}) + \left((S) \times (2 + R_{in}^2) \times R_{in}\right)\right) + (R_{in}^3 + R_{in}^2 + 1)$. Similarly total number of operations for outer sector are $\left((2 \times (R_{out} - R_{in})) + \left((S) \times (2 + (R_{out} - R_{in})^2) \times (R_{out} - R_{in})\right)\right) + ((R_{out} - R_{in})^3 + (R_{out} - R_{in})^2 + 1)$.

Now, it is trivial that for RTLNP with $R_{in} = 3$, $R_{out} = 6$ and the number of sectors $S = 10$, the number of operations for computing averages of inner and outer subsectors are $(2 \times 373)$ which is a constant. The number of comparisons required to compute the binary pattern is equal to the number of sectors $S = 10$, which is again a constant. Hence the total number of operations to encode a single pixel is constant. Therefore the overall complexity of the proposed descriptor is $O(X \times Y)$ for an image of size $X \times Y$. This complexity analysis shows that if we consider summation, comparison, multiplication and division as fundamental operations then the feature computational complexity of RTLNP is in the order of the size of the image (i.e. $O(X \times Y)$ for image of size $X \times Y$).

3. performance measures

Performance of the proposed *RTLNP* has been analyzed with respect to retrieval and recognition accuracies. Performance measures used to evaluate retrieval accuracy are Average Retrieval Precision (ARP) and Average Retrieval Rate (ARR). Precision is computed as

$$P_r(I_q, \lambda) = \frac{1}{\lambda} \sum_{i=1}^{|DS|} \Delta(\omega(I_q), \omega(I_i), \tau(I_q, I_i), \lambda) \mid I_i \neq I_q \qquad (9)$$

where $\lambda$ is the number of images retrieved, $I_q$ is the query image, $|DS|$ is the size of the dataset, $\omega(.)$ returns the class of an image and $\tau(I_q, I_i)$ is the rank of the $i^{th}$ image with respect to the query image $I_q$. Image rank is computed using similarity measure $S_{\chi^2}(.,.)$ between the $i^{th}$ image in the dataset and the query image. $\Delta(.)$ is a binary function defined as

$$\Delta(\omega(I_q), \omega(I_i), \tau(I_q, I_i), \lambda) = \begin{cases} 1, & \omega(I_q) = \omega(I_i) \text{ and} \\ & \tau(I_q, I_i) \leq \lambda \\ 0, & else \end{cases} \qquad (10)$$

Average precision per class is computed as

$$AP(C_i, \lambda) = \frac{1}{|C_i|} \sum_{q=1}^{|C_i|} P_r(I_q, \lambda) \qquad (11)$$

where $C_i$ denotes the $i^{th}$ class in the dataset and $|C_i|$ denotes the number of images in the $i^{th}$ class.

ARP is calculated over the entire dataset of $N_c$ distinct classes as

$$ARP(N_c) = \frac{1}{N_c} \sum_{i=1}^{N_c} AP(C_i, \lambda) \qquad (12)$$

Recall is defined as

$$R_e(I_q, C_i, \lambda) = \frac{1}{|C_i|} \sum_{i=1}^{|DS|} \Delta(\omega(I_q), \omega(I_i), \tau(I_q, I_i), \lambda) \mid I_i \neq I_q \qquad (13)$$

Average recall per class and ARR over the entire dataset are calculated as

$$AR_e(C_i) = \frac{1}{|C_i|} \sum_{q=1}^{|C_i|} R_e(I_q, |C_i|) \qquad (14)$$

$$ARR(N_c) = \frac{1}{N_c}\sum_{i=1}^{N_c} AR_e(C_i) \tag{15}$$

F-Score is another measure used to analyze the performance of the descriptors, which is defined as

$$F(N_c) = \frac{2 \times ARP(N_c) \times ARR(N_c)}{ARP(N_c) + ARR(N_c)} \tag{16}$$

Average normalized modified retrieval rank (ANMRR) defined in [18] is used to measure the performance of the descriptors based on the rank of the retrieved images. Low ANMRR indicates that the images retrieved by the descriptor are highly relevant to the queried image and higher value of ANMRR indicates that most of the top ranked retrieved images by the descriptors are not relevant to the queried image. The descriptor with highest F-Score should achieve least ANMRR value.

## 4. performance analysis

Performance of the proposed method has been analyzed on the latest and most challenging facial image databases namely: AT&T [19], LFW [20], Color FERET [21] [22], and CASIA-Face-V5-Cropped [23]. Lengths of the proposed and other state of the art descriptors are shown in Table 1. Length of the proposed descriptor at $\Delta\theta = 36°$ is equal to the length of LDP. The accuracy of the descriptor $RTLNP$ with different distance measure has been computed on AT&T database and shown in Fig. 8. Since the $\chi^2$ distance measure offers the best result, it has been used in all the experiments to show the robustness of the proposed descriptor against illumination, light, expression, scale, and pose variations. Four performance measures namely ARP, ARR, F-score, and ANMRR are used in all the experiments. F-score and ANMRR are computed for the maximum number of images in a particular category of the database. APR, ARR, F-Score, and ANMRR are computed by taking each image in the database as probe and rest of the images as gallery.

### 4.1. Performance analysis on AT&T database

AT&T database contains 10 different images of 40 subjects in different pose. For some the subjects images are taken with varying lighting, different expressions (open or closed eyes, smiling or non-smiling), and facial details (glasses or no-glasses). A dark homogeneous background is used for all 40 subjects.

Performance of the proposed descriptor has been analyzed to determine the appropriate $R_{in}$, $R_{out}$ and $\Delta\theta$ impirically. Experiments are conducted with varying $\Delta\theta$ and fixed values of $R_{in}$ and $R_{out}$ and the results are shown in Fig. 9. Fig. 9 shows that $RTLNP$ achieves approximately 12% better ARP and ARR with $\Delta\theta = 36°$. The highest value of F-score and the lowest value of ANMRR at $\Delta\theta = 36°$ indicate that $RTLNP$ performs best at $\Delta\theta = 36°$. Hence in the rest of the experiments that $\Delta\theta$ has been fixed to 36°.

To determine suitable $R_{in}$ and $R_{out}$, $RTLNP$ has also been analyzed with varying $R_{in}$ and $R_{out}$. The results with varying $R_{in}$ and $R_{out}$ are shown in Fig. 10. The $RTLNP$ descriptor performs best at $R_{in} = 3$ and $R_{out} = 6$. It is evident from Fig. 10, that the results at $R_{in} = 3$ and $R_{out} = 4, 5, 6$ are comparable. Hence we set the parameter $R_{in}$ to 3 and use three different values of $R_{out} = 4, 5, 6$ in rest of the experiments.

*Table 1: Length of descriptors*

| Descriptor | Length (bits) | Length (bins) |
| --- | --- | --- |
| LBP | 8 | 256 |
| SLBP | 8 | 256 |
| MBLBP | 8 | 256 |
| SOBEL_LBP | 8×2 | 256×2 |
| CSLBP | 4 | 16 |
| LTP | 8×2 | 256×2 |
| CSLTP | - | 9 |
| LDP | 8×4 | 256×4 |
| LTCoP | 8×2 | 256×2 |
| LMeP | 8×3 | 256×3 |
| LTrP | 8×13 | 256×13 |
| BRINT (r=1,q=1) | - | 144 |
| RTLNP($\Delta\theta = 36°$) | 10 | 1024 |

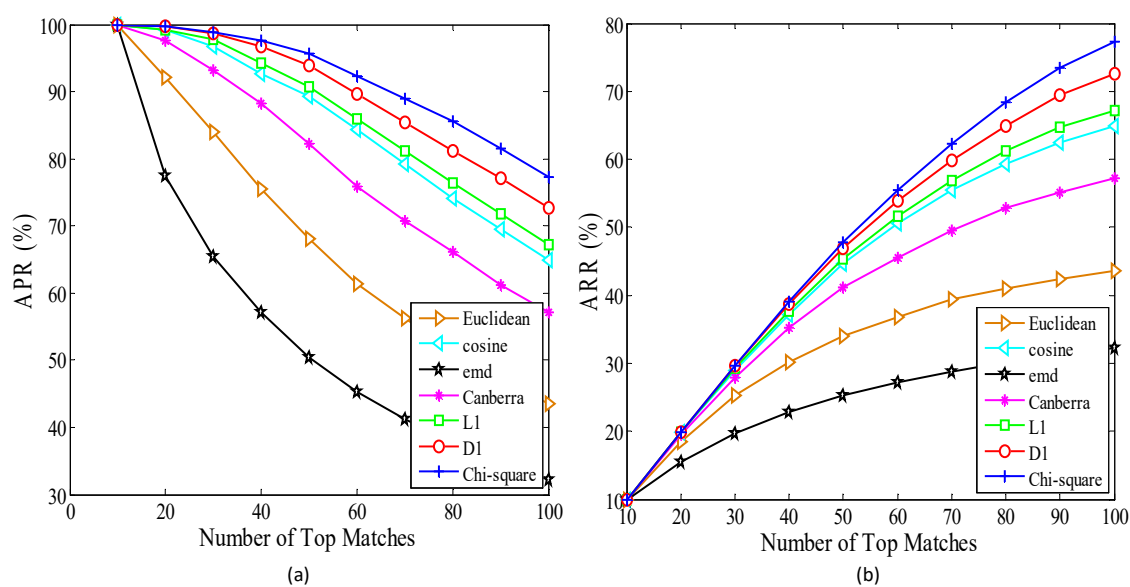

(a)     (b)

Fig. 8. Comparative analysis of RTLNP with $R_{in} = 3$, $R_{out} = 6$ and $\Delta\theta = 36°$ on AT&T with different distance measure: (a) ARP and (b) ARR.

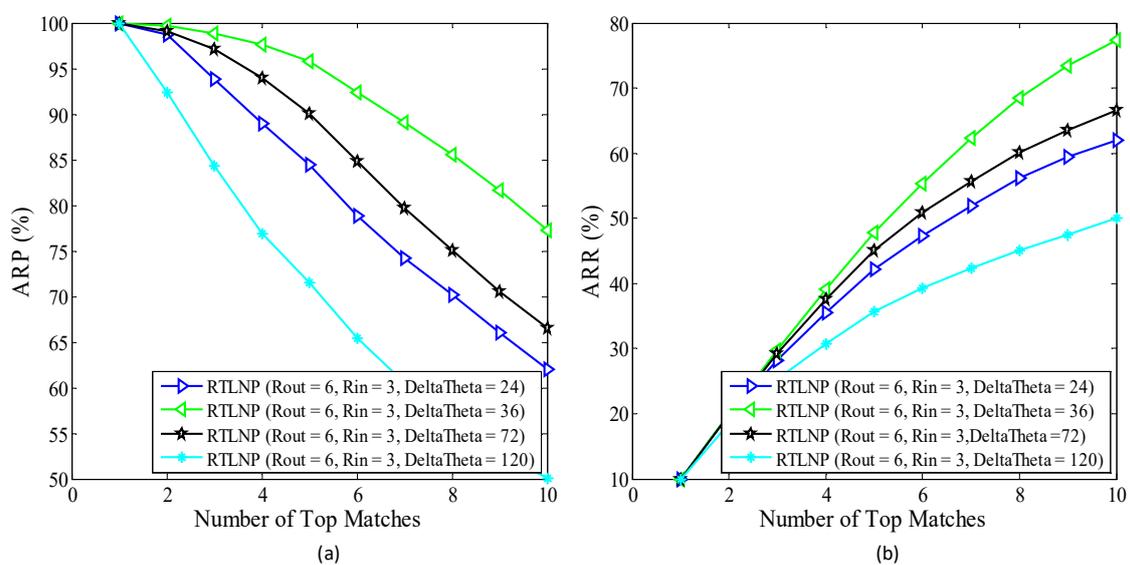

(a)                                           (b)

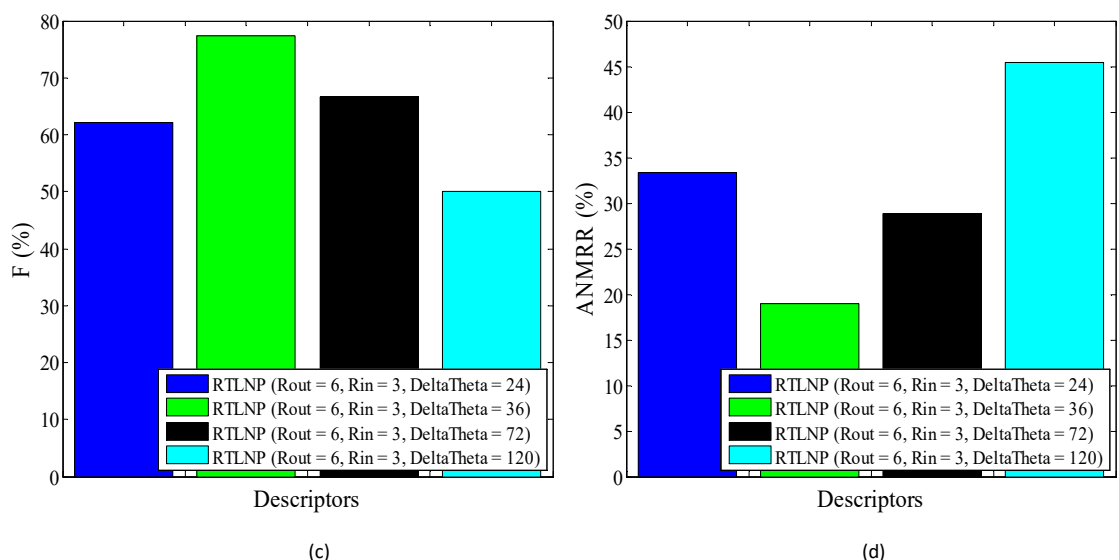

(c)                                           (d)

Fig. 9. (a) ARP, (b) ARR, (c) F-score, and (d) ANMRR computed on AT&T different $\Delta\theta$ and $R_{in} = 3$ and $R_{out} = 6$.

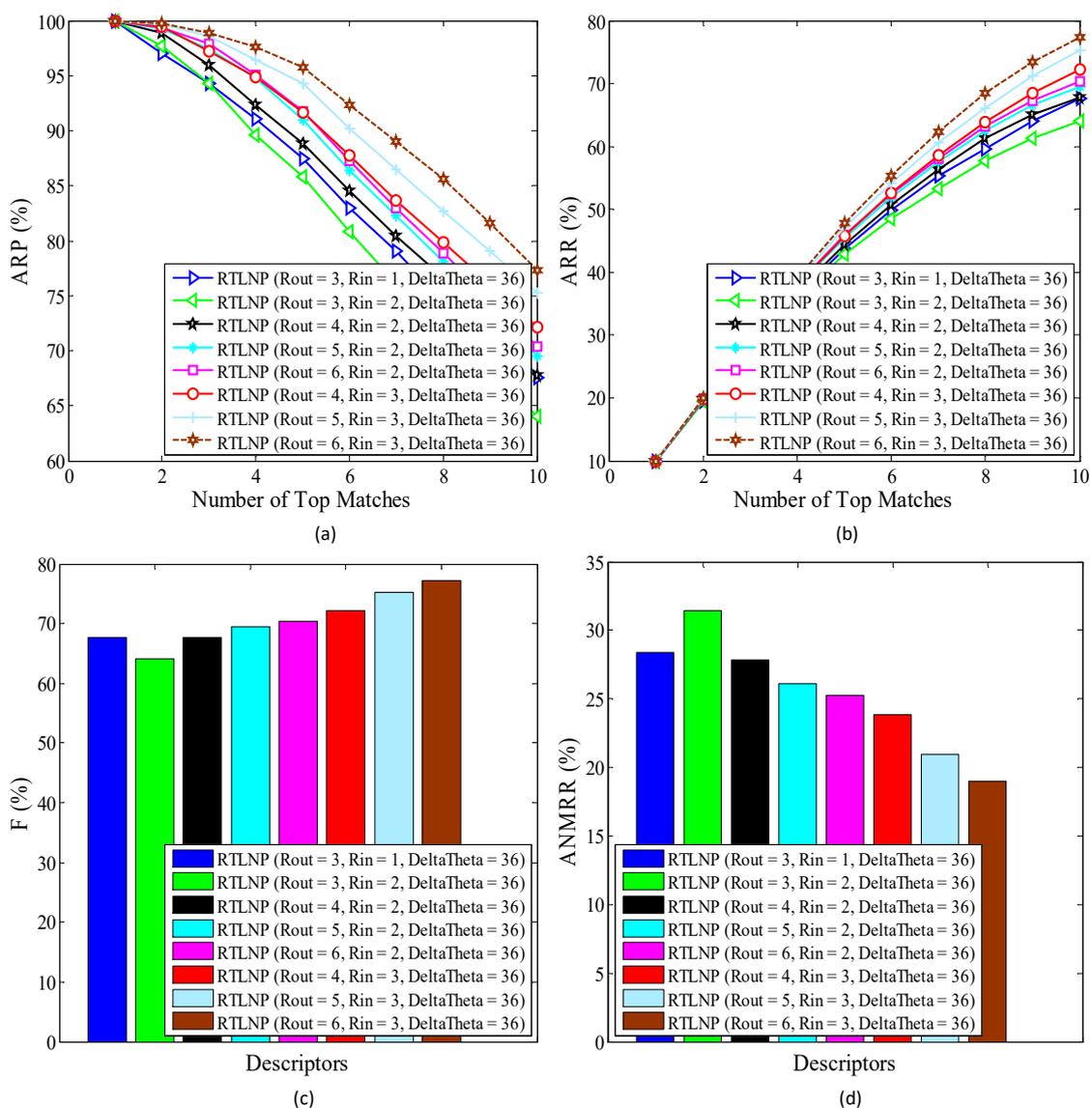

Fig. 10. (a) ARP, (b) ARR, (c) F-score, and (d) ANMRR computed on AT&T with different $R_{in}$ and $R_{out}$ and $\Delta\theta = 36°$.

### 4.2. Performance analysis on Color FERET database

"Portions of the research in this paper use the FERET database of facial images collected under the FERET program, sponsored by the DOD Counterdrug Technology Development Program Office". Color-FERET database is one of the most challenging facial image databases with severe variations in pose and expression. The color FERET database contains 11,338 facial images of 994 individuals at different orientations. There are 13 different poses used in the images of the database [21][22]. In the experiments, categories with 20 or more facial images have been used. Brief description of poses used in the facial images is given in Table 2.

The proposed descriptor is examined on color FERET database to show its capabilities against pose and expression variations. ARP and ARR values are shown in Fig. 11(a-b) for a maximum of 50 retrieved images. The ARP and ARR of RTLNP have gained approximately 10% of improvement over its nearest counterpart LTrP for 20 retrieved images, whereas, the dimension of RTLNP is approximately 0.3 times of LTrP.

F-score in Fig. 11(c) demonstrates that the proposed descriptor achieves approximately 3% improvement over its nearest counterparts. Improvement in F-score shows that the proposed descriptor mostly retrieves relevant facial images and most of the relevant images in a particular category are retrieved by the proposed descriptor. ANMRR of the proposed descriptor depicted in Fig. 11(d) is 2% lower than its nearest counterparts such as LBP, LTCoP, etc. It signifies that the proposed descriptor retrieves relevant images with lowest possible ranks (images with low rank are closer to queried image).

*Table 2: Description of different pose in images of color-feret database*

| Pose Name | Description |
|---|---|
| fa | regular frontal image |
| fb | alternative frontal image, taken shortly after the corresponding fa |
| pl | profile left |
| hl | half left - head turned about 67.5 degrees left |
| ql | quarter left - head turned about 22.5 degrees left |
| pr | profile right |
| hr | half right - head turned about 67.5 degrees right |
| qr | quarter right - head turned about 22.5 degrees right |
| ra | random image - head turned about 45 degree left |
| rb | random image - head turned about 15 degree left |
| rc | random image - head turned about 15 degree right |
| rd | random image - head turned about 45 degree right |
| re | random image - head turned about 75 degree right |

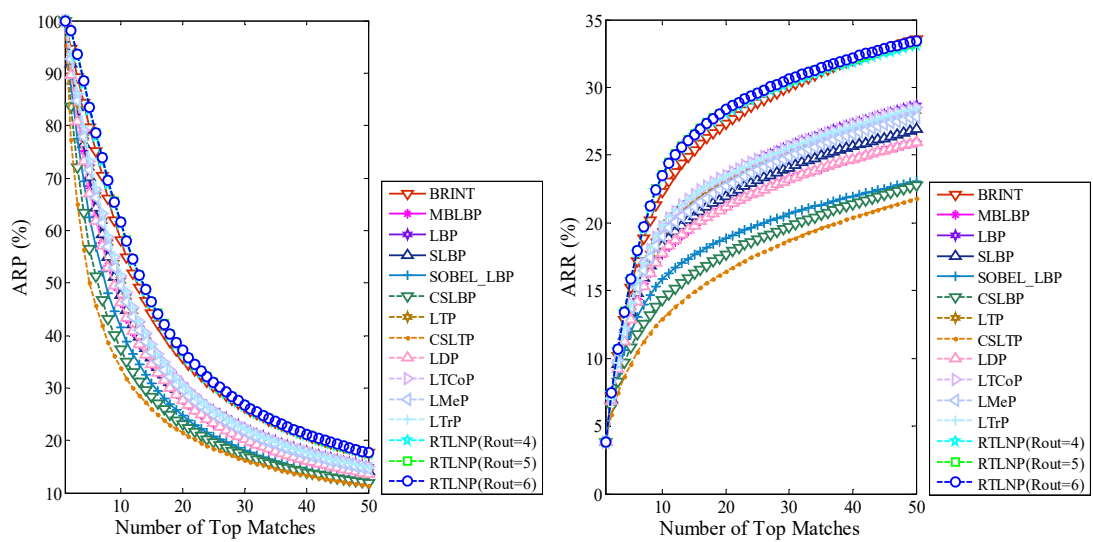

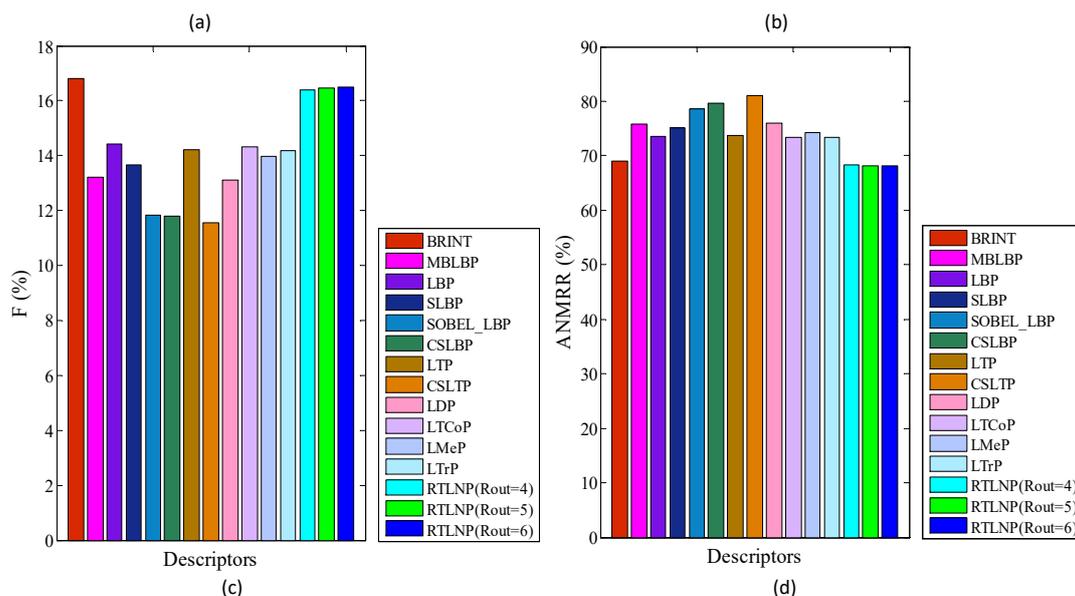

Fig. 11. (a) ARP, (b) ARR, (c) F-score, and (d) ANMRR computed on Color-FERET for different descriptors and RTLNP with $R_{in}$= 3 and $\Delta\theta = 36°$.

### 4.3. Performance analysis on LFW database

There are 13,233 color facial images of 5,749 individuals in LFW database [20]. 1680 individuals have two or more images and rest of the individuals have only one image. As the problem of unconstrained face recognition is one of the most general and fundamental face recognition problems, we test the proposed descriptors over this database. Experimental analysis is done on a subset of LFW database, i.e. images of individuals having at least 20 images are taken. Similar settings are used for experiments as used in section 4.2.

The descriptor is tested on LFW database as it is one of the largest and most challenging database of facial images taken under uncontrolled real world environment. Performance of the proposed descriptor has been evaluated on LFW to show its robustness against unconstrained variations in pose, illumination, and expression. ARP and ARR values are depicted in Fig. 12(a) and (b) respectively. Improvements in ARP and ARR illustrate that RTLNP retrieves more relevant facial images as compared to other state of the art descriptors. Higher F-score shown in Fig. 12(c) indicates that most of the images retrieved by RTLNP are relevant. Lowest value of ANMRR shown in Fig. 12(d) for RTLNP indicates that RTLNP retrieves most of the low ranked images (low rank images are closer to queried image).

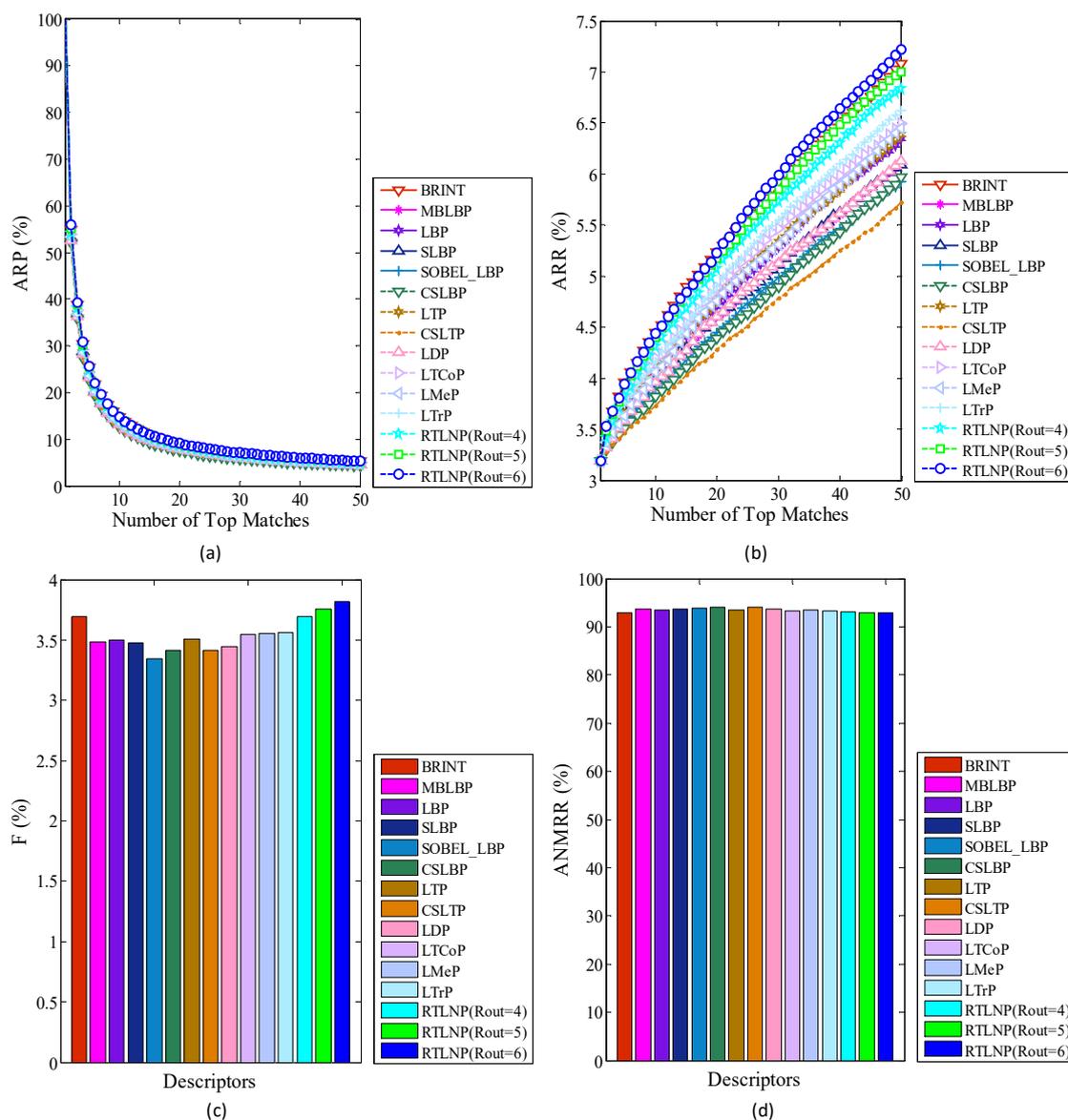

Fig. 12. (a) ARP, (b) ARR, (c) F-score, and (d) ANMRR computed on LFW for different descriptors and RTLNP with $R_{in}$ = 3 and $\Delta\theta = 36°$.

### 4.4. Performance analysis on CASIA-Face-V5-Cropped database

"Portions of the research in this paper use the CASIA-FaceV5 collected by the Chinese Academy of Sciences' Institute of Automation (CASIA)" [23]. CASIA-Face-5.0 database contains 5 color images each of the 500 individuals. Images are captured with intra-class variations such as illumination, pose, expressions, eye-glasses, and imaging distance [23].

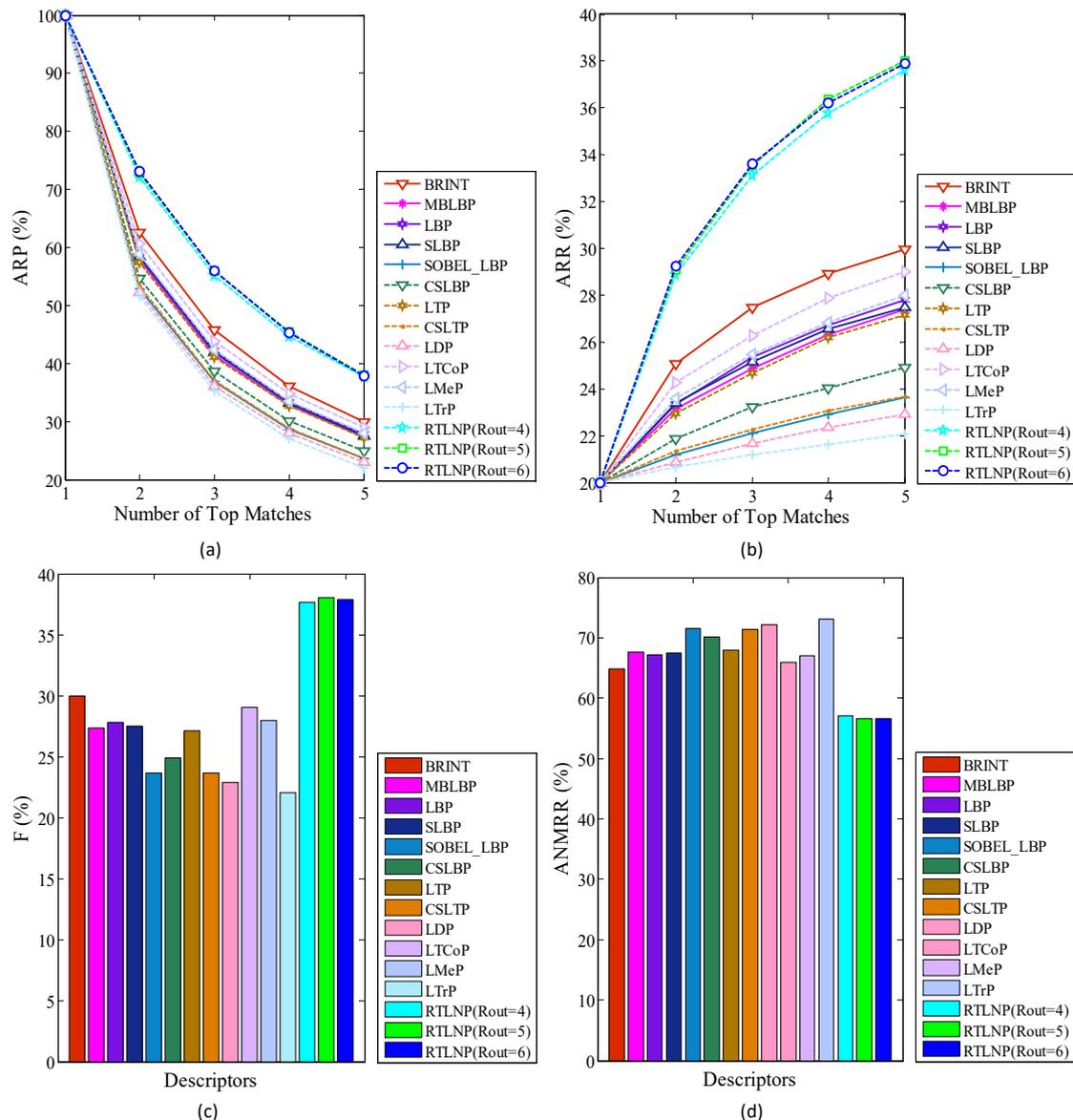

Fig. 13. (a) ARP, (b) ARR, (c) F-score, and (d) ANMRR computed on CASIA-Face-V5-Cropped for different descriptors and RTLNP with $R_{in}$= 3 and $\Delta\theta = 36°$.

Proposed Proposed descriptor RTLNP achieves better ARP and ARR over CASIA-Face-5.0 database which is confirmed through the retrieval results shown in Fig. 13(a) and (b) than most recent descriptors such as LTrP, LTCoP, LMeP, and SLBP. RTLNP achieves approximately 11% better ARP and 7% better ARR over LTCoP, 20% better ARP and 12% better ARR over LTrP, 12% better ARP and 8% better ARR over LMeP, and 11% better ARP and 8.5% better ARR compared to SLBP. It signifies that RTLNP retrieves at least 11% more relevant images than LTrP, LTCoP, LMeP, and SLBP. Higher F-score of RTLNP shown in Fig. 13(c) points out that the proposed descriptor out performs state of the art descriptors with respect to tradeoff between ARP and ARR. As shown in Fig. 13(d) RTLNP achieves lowest ANMRR, which indicates that the most of the retrieved images are having lower/better rank or non-matching images are having higher rank.

*4.5. Performance analysis in recognition framework*

Recognition rates are computed by taking each image in the database as probe and rest of the images as gallery. If there are $N$ images in the database then each image is taken as probe in turn and rest $(N-1)$ images are taken as gallery. The distance between probe feature and gallery feature is computed using $\chi^2$ distance. There are $(N-1)$ distances for each probe. The gallery image with lowest distance is given the lowest rank. If the gallery image with lowest rank belongs to the same class as the class of the probe image then it is taken as a match. The recognition rate of a descriptor is computed as

$$Recognition\ Rate = (number\ of\ matches/N) \times 100 \qquad (17)$$

The Cumulative Match Characteristics (CMCs) for different data bases are shown in Fig. 14. For CMC a match is taken, if the class of the probe image matches with the class of at least one gallery image with rank less than or equal to the maximum rank specified.

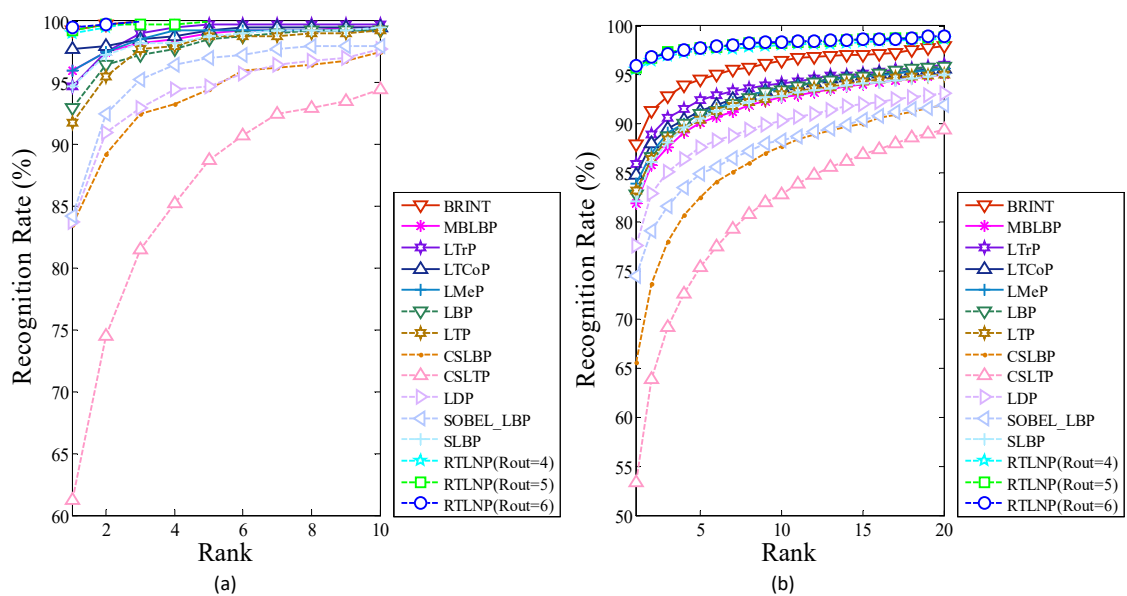

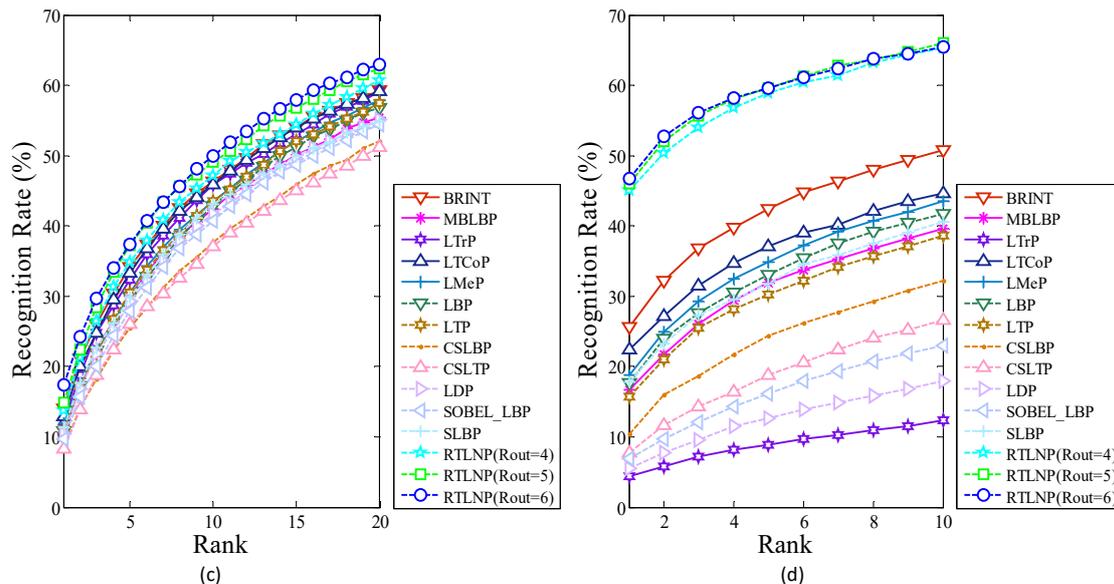

Fig. 14. CMC for different descriptors and RTLNP with $R_{in}$ = 3 and $\Delta\theta = 36°$ on databases (a) AT&T, (b) Color-FERET, (c) LFW, and (d) CASIA-Face-V5-Cropped.

The proposed method shows consistent improvement in recognition rate with increasing ranks. There is a 2% improvement in recognition rate achieved by the proposed descriptor over its nearest counterpart LTCoP on AT&T database. RTLNP shows 10% and 3% improvement over its nearest counterpart LTrP on color-FERET and LFW databases respectively. The proposed method shows significant improvement of 16% over its nearest counterpart on CASIA-Face-V5-Cropped database. Significant and steady improvement shown by the proposed method on the most challenging databases illustrates the robustness of the proposed descriptor against pose, illumination, background and expression variations.

## 5. discussions

Additional capability of the proposed descriptor on NIR facial images is explored and tested in this section. Two of the most commonly used NIR face databases CASIA NIR-VIS 2.0 [24] and PolyU-NIRFD [25] [26] have been used to examine and compare the performance of the proposed descriptor with state of the art descriptors. CASIA-NIR database contains images of 725 individuals with eye-glass, without eye-glass, expression variations, and age variations. In PolyU-NIRFD, there are images of 350 individuals with scale, pose, expression, and time (interval of 2 months) variations. The significant improvement in the retrieval results by the proposed descriptor can be observed across the ARP, ARR, F-Score and ANMRR plots of Fig. 15 as compared to the other descriptors such as LBP, LTP, SLBP, LTrP, etc. There is a 10% improvement in ARP and ARR indicating better retrieval capability of the introduced descriptor. Higher F-score and lower ANMRR values indicate the robustness of the proposed descriptors against the NIR facial imaging which is very fruitful in indoor environment.

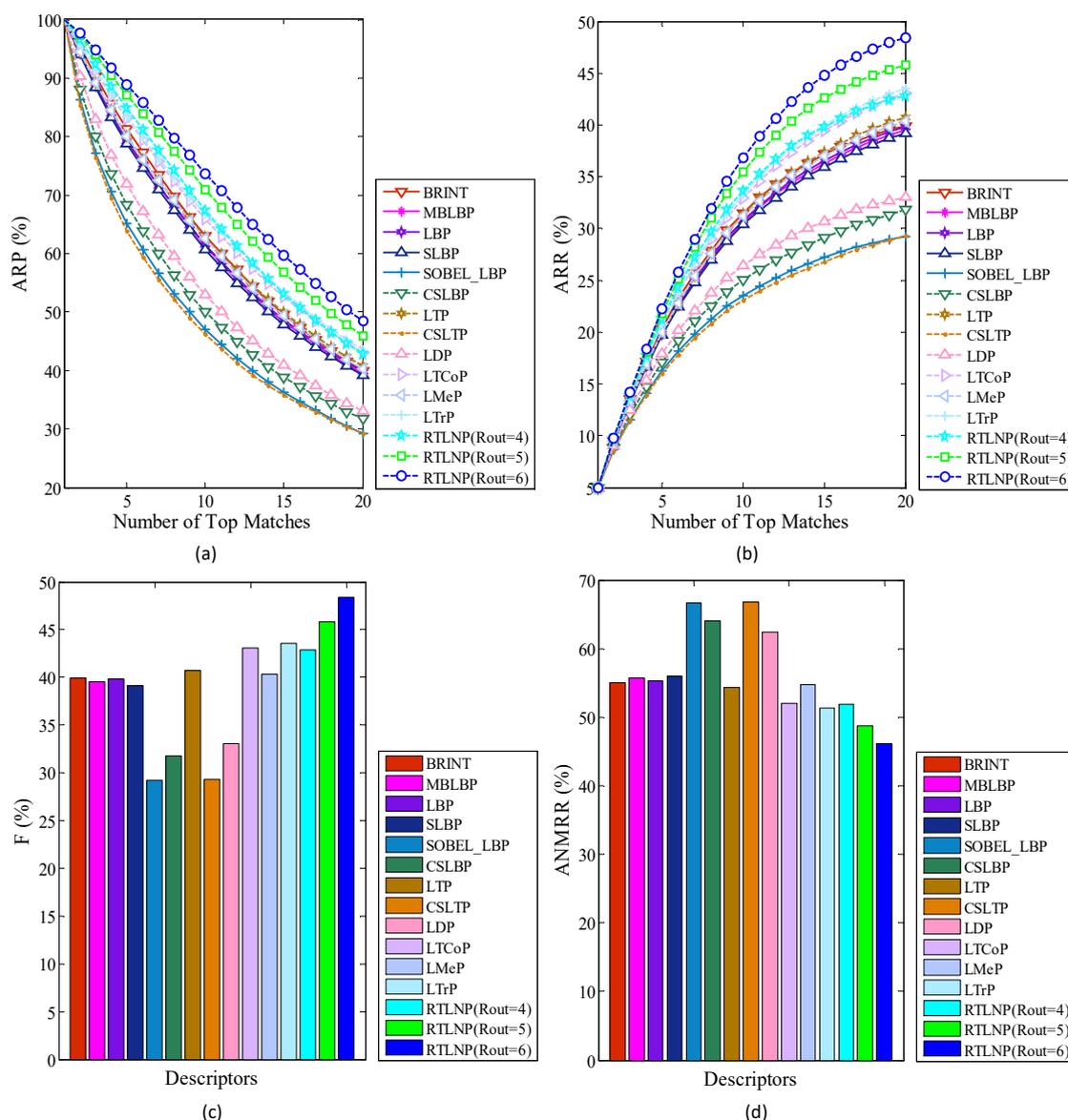

Fig. 15. (a) ARP, (b) ARR, (c) F-score, and (d) ANMRR computed on CASIA-NIR-V2.0 for different descriptors and RTLNP with $R_{in}$ = 3 and $\Delta\theta = 36°$.

Results of the experiments conducted on PolyU-NIRFD are shown in Fig. 16. RTLNP shows significant improvement with respect to number of relevant images retrieved and the quality of the retrieved images. The similar observations are also perceived for PolyU-NIR database as it was for the CASIA-NIR database. The experiments conducted over the visible as well as NIR databases suggest that the proposed descriptor can be applied for the indoor or outdoor face retrieval purpose effectively and efficiently.

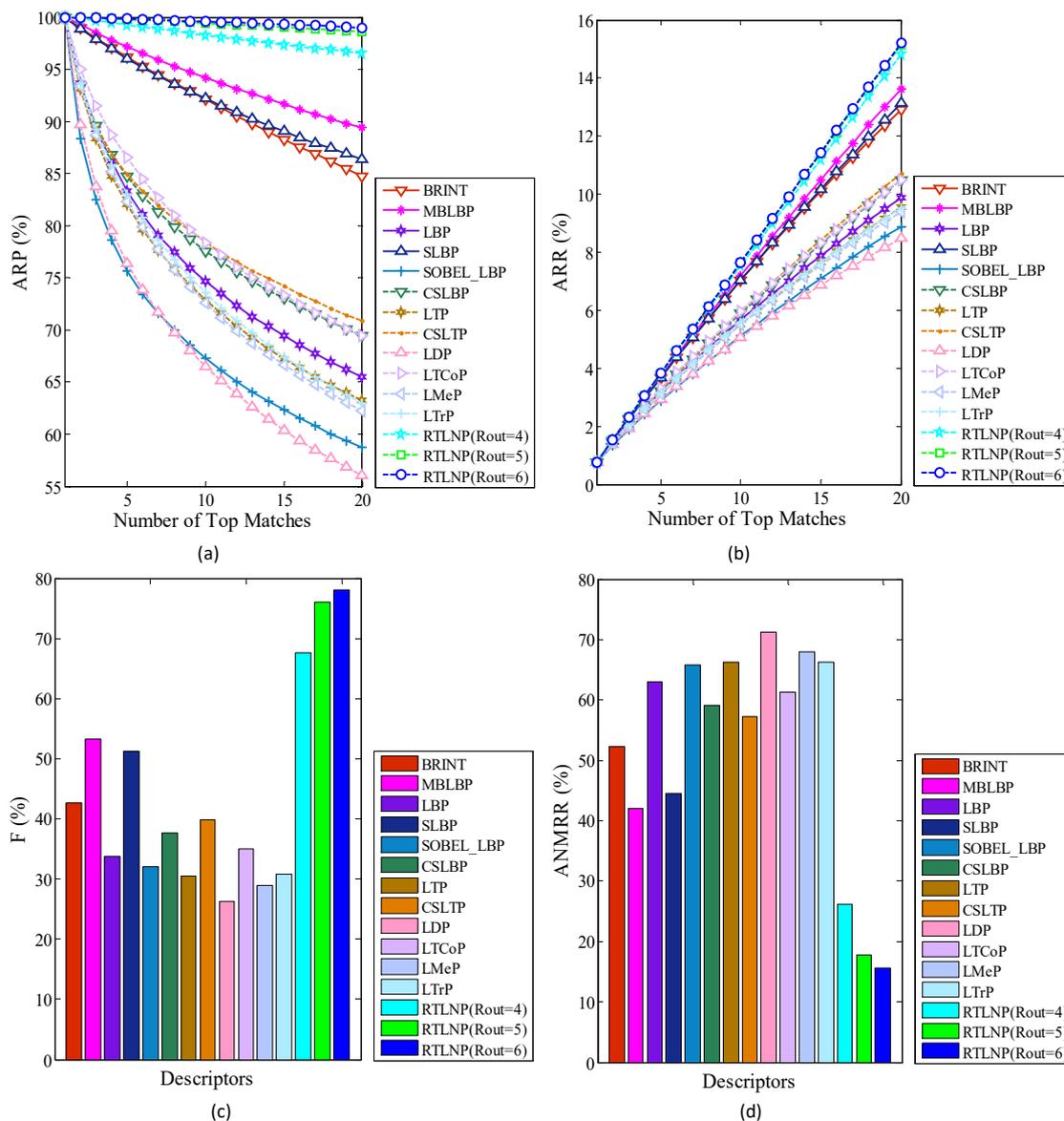

Fig. 16. (a) ARP, (b) ARR, (c) F-score, and (d) ANMRR computed on Poly-NIR for different descriptors and RTLNP with $R_{in}$ = 3 and $\Delta\theta = 36°$.

**6. Conclusion**

One of the major drawbacks of the existing descriptors was the consideration of only limited number of local neighbors. Some descriptors tried to utilize the large number of local neighbors but at the cost of curse of dimensionality. RTLNP is a local descriptor which captures distinctive relationships that exists amongst most of the pixels in the local neighborhood by expanding the neighborhood from single neighbor to a region of neighbors. As the test results show, the proposed descriptor achieves better retrieval rates than the state of the descriptors under constrained as well as unconstrained environments. Furthermore the robustness of the proposed descriptor has been confirmed through the NIR face retrieval experiments over two widely adopted databases namely CASIA-NIR and PolyU-NIR. The dimension of the descriptor does not depend upon the number of neighbors; it only depends upon the number sectors which facilitate to encode the large number of local neighbors in the descriptor construction process which in turn improves the discriminative ability of the proposed descriptor.

## 7. References

[1] P. N. Belhumeur, J. P. Hespanha, and D. J. Kriegman, "Eigenfaces vs. Fisherfaces: Recognition using class specific linear projection," IEEE Trans. Pattern Anal. Mach. Intell., vol. 19, no. 7, pp. 711–720, Jul. 1997.[r4]
[2] A. M. Martinez and A. C. Kak, "PCA versus LDA," IEEE Trans. Pattern Anal. Mach. Intell., vol. 23, no. 2, pp. 228–233, Dec. 2001.
[3] X. Xie and K.-M. Lam, "Gabor-based kernel PCA with doubly nonlinear mapping for face recognition with a single face image," IEEE Trans. Image Process., vol. 15, no. 9, pp. 2481–2492, Sep. 2006.
[4] D. Zhanga and Z. H. Zhou, "Two-directional two-dimensional PCA for efficient face representation and recognition," Neurocomputing, vol. 69, no. 1-3, pp. 224-239, 2005.
[5] H. Kong, L. Wang, E. K. Teoh, X. Li, J.G. Wang, and R. Venkateswarlu, "Generalized 2D principal component analysis for face image representation and recognition," Neural Networks, vol. 18, no. 5-6, pp. 585-594, 2005.
[6] D. Zhanga, Z. H. Zhou, and S. Chen, "Diagonal principal component analysis for face recognition," Pattern Recognition, vol. 39, no. 1, pp. 140-142, 2006.
[7] K. Etemad and R. Chellappa, "Discriminant Analysis for Recognition of Human Face Images," J. Optical Soc. Am., vol. 14, pp. 1724-1733, 1997.
[8] S. Noushath, G. Hemantha Kumar, and P. Shivakumara, "(2D)2LDA: An efficient approach for face recognition," Pattern Recognition, vol. 39, no. 7, pp. 1396-1400, 2006.
[9] S. Noushath, G. Hemantha Kumar, and P. Shivakumara, "Diagonal Fisher linear discriminant analysis for efficient face recognition," Neurocomputing, vol. 69, no. 13-15, pp. 1711–1716, 2006.
[10] T. Ojala, M. Pietikäinen, and D. Harwood, "A comparative study of texture measures with classification based on feature distributions," Pattern Recognit., vol. 29, no. 1, pp. 51–59, 1996.
[11] Marko Heikkila, Matti Pietikainen, and Cordelia Schmid, "Description of interest regions with center-symmetric local binary patterns," ICVGIP, LNCS 4338, pp. 58–69, 2006.
[12] Marko Heikkila, Matti Pietikainen, and Cordelia Schmid, "Description of interest regions with local binary patterns," Pattern Recognition, vol. 42, no. 3, pp. 425–436, 2009.
[13] Xiaoyang Tan and Bill Triggs, "Enhanced local texture feature sets for face recognition under difficult lighting conditions," IEEE Transactions on Image Processing, vol. 19, no. 6, pp. 1635–1650, Dec. 2010.
[14] Raj Gupta, Harshal Patil and Anurag Mittal, "Robust order-based methods for feature description," IEEE Conference on Computer Vision and Pattern Recognition, pp. 334–341, 2010, DOI: 10.1109/CVPR.2010.5540195.
[15] Kyungjoong Jeong, Jaesik Choi, and Gil-Jin Jang, "Semi-Local structure patterns for robust face detection," IEEE Signal Processing Letters, vol. 22, no. 9, pp. 1400-1403, 2015.
[16] B. Zhang, Y. Gao, S. Zhao, and J. Liu, "Local derivative pattern versus local binary pattern: Face recognition with higher-order local pattern descriptor," IEEE Transactions on Image Processing, vol. 19, no. 2, pp. 533–544, Feb. 2010.
[17] S. Murala, R. P. Maheshwari, and R. Balasubramanian, "Local tetra patterns: A new feature descriptor for content-based image retrieval," IEEE Transactions on Image Processing, vol. 21, no. 5, pp. 2874–2886, May 2012.
[18] Ke Lu, Ning He, Jian Xue, Jiyang Dong, and Ling Shao, "Learning view model joint relevance for 3D object retrieval," IEEE Transactions on Image Processing, vol. 24, no. 5, pp. 1449–1459, May 2015.
[19] F. Samaria and A. Harter, "Parameterisation of a stochastic model for human face identification," 2nd IEEE Workshop on Applications of Computer Vision, December 1994, Sarasota (Florida), Available: ftp://quince.cam-orl.co.uk/pub/users/fs/IEEE_workshop.ps.Z.
[20] G. B. Huang, M. Ramesh, T. Berg, and E. Learned-Miller, "Labeled faces in the wild: A database for studying face recognition in unconstrained environments," Dept. Comput. Sci., Univ. Massachusetts, Amherst, MA, USA, Tech. Rep. 07-49, Oct. 2007.
[21] P.J. Phillips, H. Wechsler, J. Huang, P. Rauss, "The FERET database and evaluation procedure for face recognition algorithms," *Image and Vision Computing J*, vol. 16, no. 5, pp. 295-306, 1998.
[22] P.J. Phillips, H. Moon, S.A. Rizvi, P.J. Rauss, "The FERET Evaluation Methodology for Face Recognition Algorithms," *IEEE Trans. Pattern Analysis and Machine Intelligence*, Vol. 22, pp.1090-1104, 2000.
[23] CASIA-FaceV5, http://biometrics.idealtest.org/
[24] Stan Z. Li, Dong Yi, Zhen Lei, Shengcai Liao, "The CASIA NIR-VIS 2.0 Face Database." In *9th IEEE Workshop on Perception Beyond the Visible Spectrum (PBVS, in conjunction with CVPR 2013)*, Portland, Oregon, June, 2013.
[25] Baochang Zhang, Lei Zhang, David Zhang, and Linlin Shen, "Directional Binary Code with Application to PolyU Near-Infrared Face Database," Pattern Recognition Letters, vol. 31, no.14, pp. 2337-2344, Oct. 2010.
[26] PolyU-NIRFD, http://www.comp.polyu.edu.hk/~biometrics/NIRFace/polyudb_face.htm.
[27] Sanqiang Zhao, Yongsheng Gao, and Baochang Zhang, "SOBEL-LBP," IEEE International Conference on Image Processing, pp. 2144–2147, 2008, DOI: 10.1109/CVPR.2010.5540195.
[28] Subrahmanyam Murala and Q.M.Jonathan Wu, "Local ternary co-occurrence patterns: A new feature descriptor for MRI and CT image retrieval," Neurocomputing, vol. 119, pp. 399–412, 2013.
[29] Subrahmanyam Murala and Q.M.Jonathan Wu, "Local mesh patterns versus local binary patterns: biomedical image indexing and retrieval," IEEE Journal of Biomedical and Health Informatics, vol. 18, no. 3, pp. 929–938, May 2014.
[30] L. Liu, Y. Long, P. W. Fieguth, S. Lao, and G. Zhao, "BRINT: binary rotation invariant and noise tolerant texture classification." *IEEE Transactions on Image Processing*, vol. 23, no. 7, pp. 3071-3084, 2014.

[31] L. Zhang, R. Chu, S. Xiang, S. Liao, & S. Z. Li, "Face detection based on Multi-Block LBP representation," In Advances in biometrics, LNCS, Springer Berlin Heidelberg, vol. 4642, pp. 11-18, 2007.
[32] Y. Taigman, M. Yang, M. A. Ranzato, & L. Wolf, "Deepface: Closing the gap to human-level performance in face verification," In Proceedings of the IEEE Conference on Computer Vision and Pattern Recognition, pp. 1701-1708, 2014.
[33] O. M. Parkhi, A. Vedaldi, & A. Zisserman, "Deep face recognition," Proceedings of the British Machine Vision, 1(3), 6, 2015.